\documentclass[a4paper,fleqn]{cas-dc}

\usepackage[numbers]{natbib}

\usepackage{lineno, hyperref}
\usepackage{graphicx} 
\usepackage[farskip=0pt]{subfig} %

\usepackage{amsmath} 

\usepackage{multirow} 
\usepackage{amssymb} %
\usepackage{pifont} 
\newcommand{\cmark}{\ding{51}} %
\newcommand{\xmark}{\ding{55}} %
\usepackage{tabularx}

\usepackage[ruled,vlined]{algorithm2e} 
\usepackage{algpseudocode} %

\SetKwRepeat{Do}{do}{while}

\def\tsc#1{\csdef{#1}{\textsc{\lowercase{#1}}\xspace}}
\tsc{WGM}
\tsc{QE}
\tsc{EP}
\tsc{PMS}
\tsc{BEC}
\tsc{DE}

\begin{document}
\let\WriteBookmarks\relax
\def\floatpagepagefraction{1}
\def\textpagefraction{.001}
\shorttitle{SiamPolar}
\shortauthors{Yaochen Li et~al.}

\title [mode = title]{SiamPolar: Semi-supervised Realtime Video Object Segmentation with Polar Representation}                      
\tnotemark[1]

\tnotetext[1]{This work is supported by National Natural Science Foundation of China under Grant no. 61803298 and 61973245.}

\author[1]{Yaochen Li}
\author[1]{Yuhui Hong}
\author[1]{Yonghong Song}
\cormark[1]
\author[1]{Chao Zhu}
\author[1]{Ying Zhang}
\author[2]{Ruihao Wang}

\address[1]{School of Software Engineering, Xi’an Jiaotong University, China}
\address[2]{MEGVII Technology, China}

\cortext[cor1]{Corresponding author}

\begin{abstract}
Video object segmentation (VOS) is an essential part of autonomous vehicle navigation. 
The real time speed is very important for the autonomous vehicle algorithms along with the accuracy metric.
In this paper, we propose a semi-supervised real-time method based on the Siamese network using a new polar representation. 
The input of bounding boxes are initialized rather than the object masks, which are applied to the video object detection tasks.
The polar representation could reduce the parameters for encoding masks with subtle accuracy loss, so that the algorithm speed can be improved significantly. 
An asymmetric siamese network is also developed to extract the features from different spatial scales. 
Moreover, the peeling convolution is proposed to reduce the antagonism among the branches of the polar head. 
The repeated cross-correlation and semi-FPN are designed based on this idea. 
The experimental results on the DAVIS-2016 dataset and other public datasets demonstrate the effectiveness of the proposed method.
\end{abstract}


\maketitle

\section{Introduction}\label{intro}

As a fundamental task in computer vision community, video object segmentation (VOS) has a wide range of application prospects in the communities of video summarization, high-definition video compression, and autonomous vehicles. 
In real scenes, video object segmentation not only faces difficulties of occlusion and deformation, but also needs to speed up to meet the demands of practical applications. Besides,  it is essentially important to segment and track online for many applications.
The word \textit{online} means during tracking only a few previous frames are utilized for inference at any time instance. 

According to whether the masks are labeled in the first frame, the VOS methods can be categorized into semi-super-\\vised and unsupervised types. 
For the semi-supervised VOS method, the masks of the objects in the first frame are manually initialized, and  the objects are tracked and segmented in the entire video. 
On the other hand, unsupervised methods try to automatically locate the objects and then segment them. 
For both of the types, the pixel-wise masks limit the segmentation speed to a certain degree. 

Inspired by the work of PolarMask \cite{xie2020polarmask}, a mask can be represented in the polar coordinate system which includes a center point and the rays emitting from the center point to edge points. 
This model simplifies the mask encoding  without obvious accuracy loss. 
In practice, it is worth sacrificing tiny accuracy for speed benefit. 
The parameter number is greatly reduced in the network using the polar representation, and a faster video segmentation speed can be achieved. 

In the Siamese network structure, the backbones are the essential part for feature extraction. Most Siamese networks use shallow network structure. For instance,  SiamFC \cite{bertinetto2016fully}, SiamRPN \cite{li2018high}, SiamRPN++ \cite{li2019siamrpn++} and SiamMask \cite{wang2019fast} all use the backbones of AlexNet and ResNet50. The asymmetric backbones not only introduce deeper backbones into Siamese networks, but also balance the speed and accuracy of video segmentation. 

Antagonism problem \cite{peng2017large, li2019siamrpn++, zhang2019deeper} exists among the different tasks in video object segmentation. In fact, the tasks of network branches are usually contradictory. For the tasks of classification, center location and tracking, the models are required to be invariant to various transformations such as translation and rotation. However, for the segmentation task, the models should be transformation-sensitive to precisely specify the semantic category for each pixel.
The idea of peeling convolutions in the network is proposed to release the feature differentiation without increasing the parameters. In this way, the antagonistic effects among branches are weakened. Peeling convolution can be used as a  criterion for all the Siamese-network-based object tracking and segmentation methods. Based on the idea of peeling convolution, a novel correlation computation method based on feature pyramid network is designed in this paper. 

The main contributions of this paper are summarized as follows: 

\begin{itemize}
  \item Improved polar representation: We introduce a novel polar representation for video object segmentation and propose a real-time video object segmentation method named SiamPolar, which is also an anchor-free object tracking method.

  \item Asymmetric Siamese network: An asymmetric Siam-\\ese network is developed using similar backbones with different depths, which not only get a more accurate correlation similarity by aligning the features, but also allows the siamese network to perform better in video segmentation with deeper backbones. 

  \item Peeling convolutions: Negative effects exist among the branches of polar head, so we design repeated cross correlation and semi-FPN based on the idea of peeling convolutions. Redundant anti-features can be reduced without convolutions. As a result, the mutual influence between each branch feature can be decreased.
\end{itemize}

The rest of the paper is summarized as follows: Section II shows an overview of the related works. In Section III and IV, the proposed methodology is introduced. The experiments are conducted in Section V. Finally, we close this paper with conclusions in Section VI.

\section{Related Work}

\begin{figure}[tpb]
	\centering
	\includegraphics[width=\linewidth]{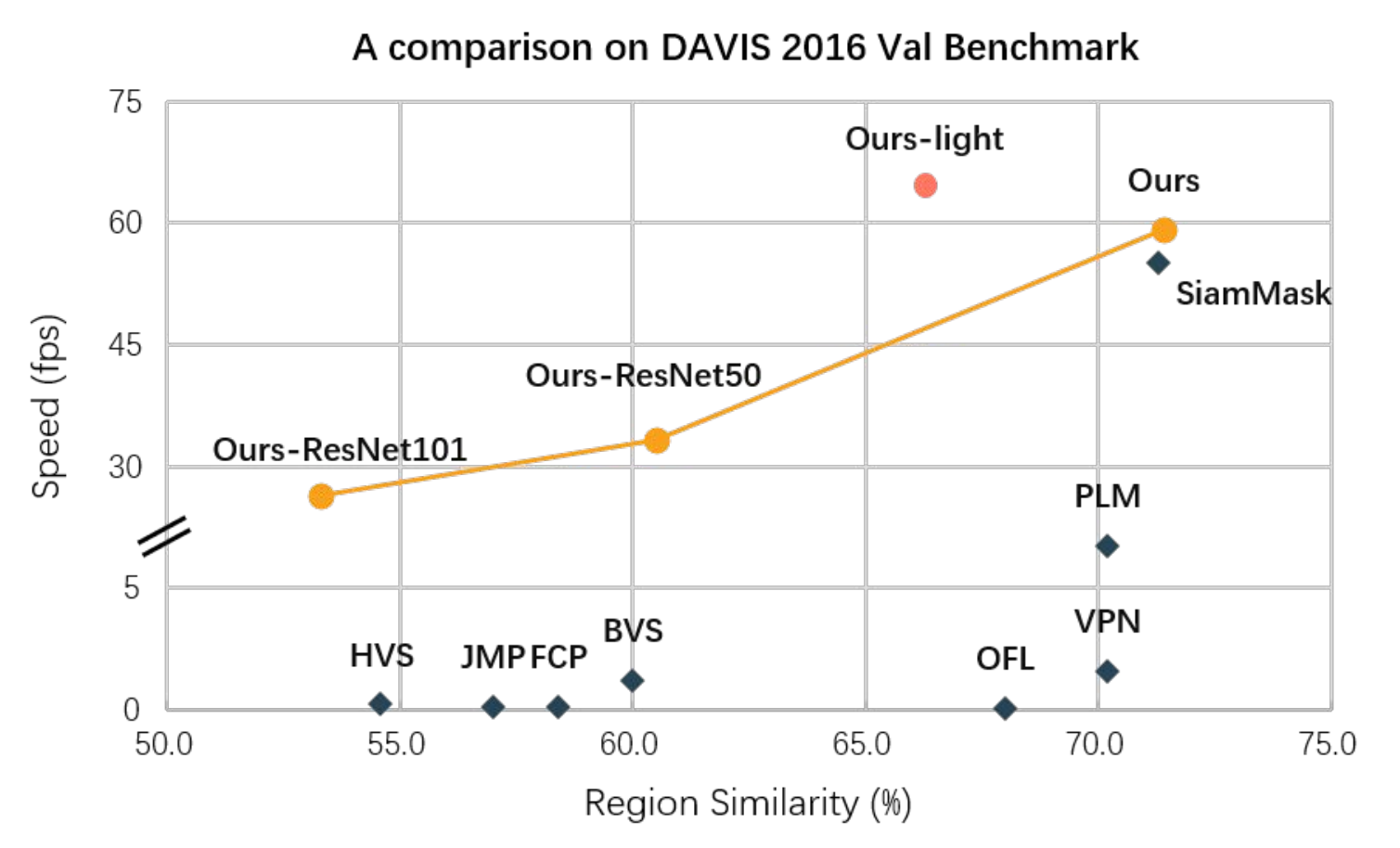}
	\caption{Comparison of the quality (region similarity) and the run-time (frame per second) of the methods on the DAVIS 2016 validation benchmark. 'Ours' denotes our full network with asymmetric backbones and semi-FPN. 'Ours-ResNet50' ('Ours-ResNet101') utilizes semi-FPN and two ResNet50 (ResNet101) as backbones. 'Ours-light' does not use FPN nor semi-FPN. }
	\label{fig2: comparison}
\end{figure} 

In the traditional computer vision methods, video segmentation is regarded as a spatiotemporal label propagation problem. These methods usually rely on two important cues: object representation of graph structure and spatiotemporal connections. Commonly used object representation methods include pixels, superpixels, and object patches. The graph structures are then constructed (such as conditional random field, Markov random field \cite{bao2018cnn}, or a mixture of trees) based on these representations. On the other hand, establishing a connection between time and space is also an important aspect to improve the accuracy of video object segmentation, which can usually be done through spatiotemporal lattices \cite{marki2016bilateral}, nearest neighbors \cite{fan2015jumpcut}, and hybrid trees \cite{badrinarayanan2013semi} to establish connections between frames. Among them, bilateral space video segmentation \cite{marki2016bilateral} is a method of video segmentation in bilateral space. They design a new energy function on the vertices of the space-time bilateral grid and then use standard graph cuts to complete video segmentation. 
With the development of neural networks, its advantages in video segmentation have gradually emerged. These methods can be divided into two categories including: (1) motion-based video segmentation and (2) detection-based video segmentation. 

The motion-based methods take advantage of the continuity of object motion. The fully connected object proposals \cite{perazzi2015fully} proposed by Federico et al. establishes an energy function on the fully connected graph of the candidate object, and then maximizes this energy function by support vector machine to distinguish the foreground and the background obtaining the segmentation result. Video propagation networks \cite{jampani2017video} propose a video propagation technology through a temporal bilateral network for dense and video adaptive filtering to get the segmentation results. 
Moreover, the continuity of video motion is usually expressed as optical flows. The OFL method \cite{perazzi2015fully} proposed by Yi-Hsuan et al. establishes a multi-level spatial-temporal model, which recalculates the optical flow at the segmentation edge to obtain the object flow to fine-tune the results. The recurrent neural network (RNN) can also be applied to utilize motion information. For instance, MaskRNN \cite{hu2017maskrnn} constructs an RNN and fuses the binary mask and the output of the positioning network with the optical flow in each frame. Recently, \citet{bao2018cnn} propose an accurate method that uses spatiotemporal Markov random field, in which temporal correlation is modeled by optical flow, and spatial correlation is represented by convolutional neural network (CNN). 

The detection-based video segmentation methods train the appearance model and conduct object detection and segmentation for each frame. The PLM method \cite{shin2017pixel} is a CNN-based pixel-level method, which distinguishes the object region from the background based on the pixel-level similarity between two objects. \citet{caelles2017one} use offline and online training of fully convolutional neural networks for one-shot video object segmentation. 
This pre-trained network can be fine-tuned on the first frame of the video. Besides, they extend the model with explicit semantic information, which significantly improve the accuracy of the results \cite{maninis2018video}. To achieve higher accuracy, the VOS methods usually adopt computationally intensive strategies. Therefore, these methods  show low frame rates and cannot show real-time performance.

Furthermore, researchers get interested with faster methods such as \cite{marki2016bilateral, wug2018fast, chen2018blazingly, cheng2018fast, jampani2017video, yang2018efficient, nekrasov2019real}. Among them, the most effective methods are based on Yang et al. \cite{yang2018efficient} and Wug et al. \cite{wug2018fast}. The former method utilizes a meta-network “modulator” to make the parameters of the segmentation network converge quickly, while the latter uses a multi-stage encoder-decoder structure suitable for Siamese networks, both of \\which run below 10 fps. 
Recently, the Siamese networks perform well for the task of object tracking. Qiang et al. \cite{wang2019fast} first introduce Siamese network to the task of video segmentation unifying object tracking and segmentation. The SiamMask method greatly improves the speed of VOS. Moreover, only the initialization of bounding box is needed rather than the mask. 
The proposed method draws on the idea of SiamMask but achieves faster speed and higher accuracy. The comparison with the state-of-the-art methods is shown in Figure~\ref{fig2: comparison}. 

\section{Improved Polar Representation}\label{improve_polar_representation}

The polar representation is first introduced by Xie et al. as an instance segmentation method named PolarMask \cite{xie2020polarmask}. It shows a much faster speed than other methods, which is crucially important for unmanned vehicles. 
However, the polar representation performs poor by only keeping the largest contour, especially when the object is occluded.
 It can be improved by merging the contours and modifying the centers. 

\subsection{Label Generation} 

\begin{figure*}
	\centering
	\includegraphics[width=\linewidth]{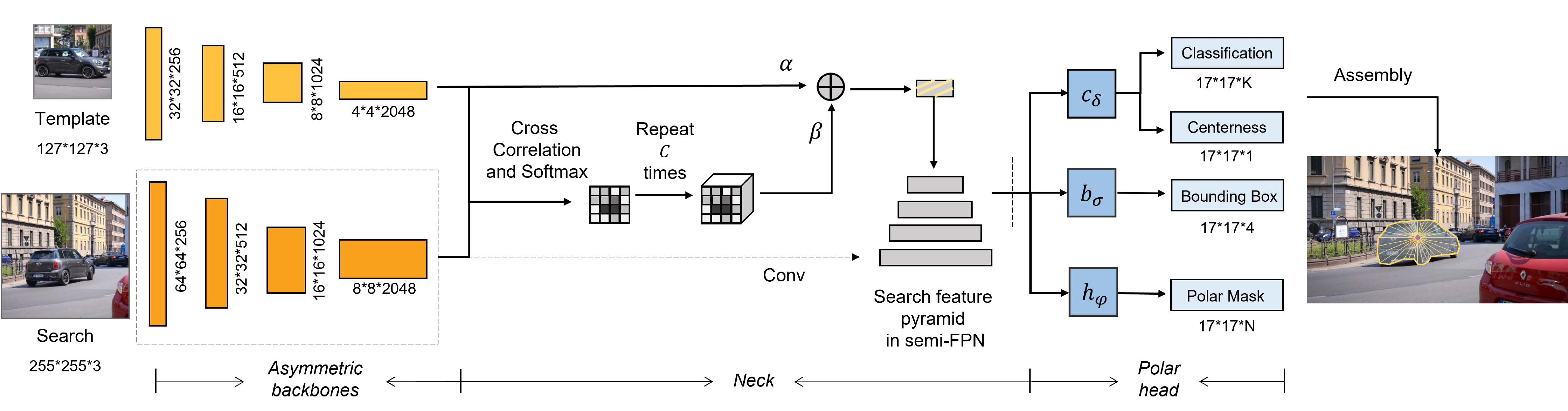}
	\caption{The Network structure of SiamPolar consists of three components: asymmetric siamese backbone, neck, and polar head. The detailed structure of semi-FPN is shown in Figure~\ref{fig7: semi_fpn}. The dashed line between the neck and the polar head indicates that the polar head is repeated for each layer of the search feature pyramid. $N$, $K$, $C$ denote the number of rays, the number of categories and the number of channels. $\alpha$ and $\beta$ represent the coefficients of the  features for feature fusion. $c_{\delta}$, $b_\sigma$ and $h_{\phi}$  denote the stacked convolutions to extract different features for classification, candidate centers, bounding box and polar represented mask.}
	\label{fig3: structure_SiamPolar}
\end{figure*}

In the polar coordinate system, the mask of any instance can be represented by a candidate center and the rays emitting from the center to mask edge. To simplify this computation, the angular intervals between rays are specified equally. For video object segmentation, we only need to predict the center coordinate $(x_c, y_c)$ and the distance $\{d_1, d_2, ..., d_n\}$ \\from each point to the candidate center. The assemble results are shown in Figure~\ref{fig3: structure_SiamPolar} (right). This polar representation method reduces the dimensionality of the mask representation. Therefore, it leads to a substantial speed up as it reduces the parameter numbers of the mask prediction branch. 

Moreover, two special cases should be emphasized: (1) If a ray has multiple intersections with the contour, we select the one with the largest length. (2) For each ray lying angle $\theta$, we set the distance value to the minimum if there is no contour point all in $\theta_{near} = \theta \pm \delta$.
The ray will be filtered out as a negative sample in the non-maximum suppression. 

To generate a mask represented by polar coordinates, we first locate the object contours using OpenCV Toolbox. 
The object contours will be merged by Algorithm~\ref{alg:B} if the object is occluded.
The diameter of the contours $C$ is defined as: 
\begin{equation}
	\begin{aligned}
		W(C) = \sqrt{\text{max length}^2(C) + \text{max width}^2(C)}
	\end{aligned}
\end{equation}
This method will bring some background pixels into the generated masks, however, it can  cover a bigger area of masks. The merge ratio $\mu$ can be adjusted in the experiments. Then Algorithm~\ref{alg:A} is applied to generate the distance labels. 
We define that all the masks are generated by 36 rays.
The  effect of ray number will be discussed in Section~\ref{ray_number_influence}. 

\begin{algorithm}
    \caption{Contours Merger \label{alg:B}}
    \DontPrintSemicolon
    \KwData{Contours: $Contours$}
    \KwResult{Merged Contours: $R$}
    \Begin{
        Find the biggest contour $b$\;
        Initiate $R \gets [b]$\;
        $p \gets$ midpoint of $b$\;
        \Do{
            No contour is added\;
        }{
            \For{each contour $c_i \in Contours$}{
                $p_i \gets$ midpoint of $c_i$\;
                \If{distance between $p$ and $p_i$ is less than $\mu (W(R) + W([c_i]))$}{
                    Append $C_i$ into $R$\;
                    $p \gets$ midpoint of $p$ and $p_i$\;
                }
            }
        }
        \KwRet{$R$}
    }
\end{algorithm}

\begin{algorithm}
    \caption{Distance Label Generation (e.g.,36 rays)\label{alg:A}}
    \DontPrintSemicolon
    \KwData{Contours: $Contours$, Center Sample: $center$}
    \KwResult{Distance Label: $L_D$}
    \Begin{
        Initialize distance set D, angle set A\;
        \For(){each $contour \in Contours$}{
            \For{each $point \in contour$}{
                Calculate distance and angle from $point$ to $center$\;
                Append distance to D, angle to A\;
            }
        }

        Initialize distance label $L_D$\;
        \For{angle $\theta \in [10, 20, ..., 360]$}{
            \uIf{Find angle $\theta$ in A}{
                \lIf{angle has multiple distances $d$}{Get the maximum $d$}
                \lElse{Get the corresponding distance $d$}
            }
            \uElseIf{Find angle $\theta_{near} = \theta \pm \delta$ in A}{
                Get corresponding $d$
            }
            \Else{
                $d \gets 10^{-6}$
            }
            Append $d$ to $L_D$\;
        }
        \KwRet{$L_D$}
    }
\end{algorithm}

\subsection{Candidate Center}

The selection of centers will not affect the accuracy of the generated masks, however, a better center can generate a group of relatively consistent line segment length.
The rays with similar line segment length can be easily trained for neural networks.
According to data smoothing method of \cite{simonoff2012smoothing}, the smooth data can identify simplified changes to help predict different trends and patterns. 

The best object center can be selected from a group of candidate centers by sampling around the mass center.
Firstly, the regions with a certain $stride$ \cite{tian2019fcos} around the mass center are examined.
If the location $(x, y)$ falls into this region, it will be considered as a positive sample. Otherwise, it is deemed as a negative sample. The number of positive samples is expanded from $1$ to $9$ (or $16$), which can balance the positive and negative samples. 

\begin{figure}[tpb]
	\centering
	\includegraphics[width=\linewidth]{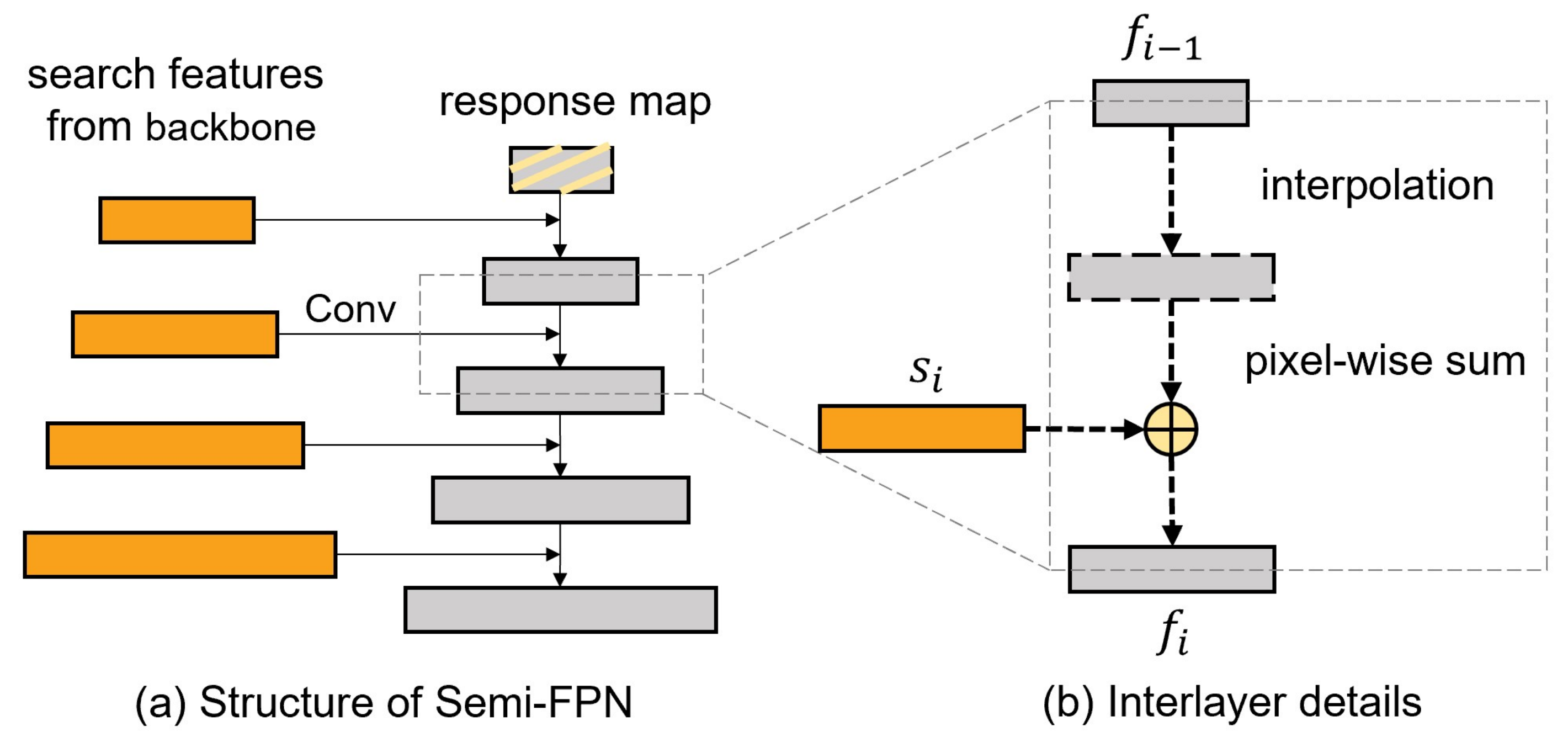}
	\caption{The detailed structure of semi-FPN. (a) The basic structure of Semi-FPN. (b) The detailed feature fusion method of Figure (a). $f_{i-1}$ and $f_i$ denote the features of FPN, while $s_i$ denotes the feature of search backbone.}
	\label{fig7: semi_fpn}
\end{figure} 

\subsection{Loss Functions}

Focal Loss \cite{lin2017focal}, smooth $L_1$ loss \cite{girshick2014rich} and Polar IoU Loss are utilized as the loss functions of classification, bounding box and polar mask. 
\begin{align}
	\begin{split}
		\mathcal{L}_{cls}=-(1-p_{pred} \times p_{gt})^\gamma \log(p_{pred})
	\end{split}
\end{align}
\begin{align}
	\mathcal{L}_{bbox}=\sum_{i \in [t,b,l,r]}
		\left\{
		\begin{aligned}
		0.5 \times (d^i)^2, & \text{ if }|d^i| \leq 1\\
		|d^i| - 0.5, & \text{ if }|d^i|>1
		\end{aligned}
	\right.
\end{align}
where $\gamma$ denotes an experimental value, $d^i$ is the distance between center and edge of the bounding box ($d^i=b^i_{pred}-b^i_{gt}$). $t$, $b$, $l$, $r$ denote top, bottom, left and right, respectively. 

Polar Centerness is designed to down-weight the high diversity of rays’ lengths in distance regression, which makes the network choosing the best center candidate. The original Polar Centerness only utilizes the maximum and minimum statistics. It can not get the best trainable center candidate in extreme situations where the distances are distributed around the maximum and minimum values. 
For instance, two groups of rays are generated from the same mask with the lengths of $(1,2,3)$ and $(1,3,3)$ respectively. 
For training, the pattern of $(1,2,3)$ can be recognized by the network more easily than $(1,3,3)$, however, the original Polar Centerness cannot prove that. 
Our improved Polar Centerness can figure out that the pattern of $(1,2,3)$ outperforms $(1,3,3)$. 
The average values are computed to solve this problem. 
The original and the new Polar Centerness are defined as follows: 
\begin{equation}
	\begin{aligned}
		original&\;Polar\;Centerness \\
		& = \sqrt{\frac{min(d_1, d_2, ..., d_n)}{max(d_1, d_2, ..., d_n)}}
	\end{aligned}
	\end{equation}
\begin{equation}
\begin{aligned}
	\mathcal{C}_{polar} 
	& = improved\;Polar\;Centerness \\
	& = \sqrt{\frac{1}{2} \times \left[ \frac{min(d_1, d_2, ..., d_n)}{mean(d_1, d_2, ..., d_n)} + \frac{mean(d_1, d_2, ..., d_n)}{max(d_1, d_2, ..., d_n)} \right]}
\end{aligned}
\end{equation}

In most cases, object detection and segmentation methods utilize smooth $L_1$ loss or IoU loss. The definition of IoU is the ratio of interaction area over union area between the predicted and the ground-truth masks. In polar correlation, IoU can be defined as:

\begin{equation}
	IoU=\frac{\int_{0}^{2\pi}\frac{1}{2}\min(d, d^*)^2d\theta}{\int_{0}^{2\pi}\frac{1}{2}\max(d, d^*)^2d\theta}
\end{equation}
where the regression target $d$ and the predicted $d^*$ are the target and predicted ray length at the angle $\theta$. To use the polar representation method introduced above, the IoU metric can be expressed as follows:

\begin{equation}
	IoU=\lim_{n\rightarrow\infty}\frac{{\sum_{i=1}^N}\frac12d_{min}^2\Delta\theta_i}{\sum_{i=1}^N\frac12d_{max}^2\Delta\theta_i}
\end{equation}
where $d_{min} = \min(d, d^*)$ and $d_{max} = \max(d, d^*)$. If $N$ is a finite value, $\triangle\theta=\frac{2\pi}N$. As the square operation has little influence on the accuracy, the above equation can be further simplified. The binary cross-entropy of polar IoU is then computed to get the loss function: 

\begin{equation}
	\mathcal{L}_{mask}=\log\frac{\sum_{i=1}^Nd_{max}}{\sum_{i=1}^Nd_{min}}
\end{equation}

During the inference, we multiply Polar Centerness with classification to obtain the final confidence scores. Thus, the final loss function is defined as:
\begin{equation}
	\mathcal{L}=\mathcal{L}_{cls} \times \mathcal{C}_{polar} +\mathcal{L}_{bbox}+\mathcal{L}_{mask}
\end{equation}

The masks with top-scoring predictions per FPN level are assembled. The redundant masks are removed by non-maximum suppression (NMS). Then it is easy to assemble the contour points into the mask. The point at each fixed angle can be computed by the following equation: 
\begin{align}
	x_i = \cos\theta_i \times d_i + x_c \\
	y_i = \sin\theta_i \times d_i + y_c
\end{align}
where $(x_c, y_c)$ denotes the center, $\theta_i$ denotes the ray angle, and $d_i$ denotes the $i_{th}$ ray length.

\section{Network Structure}

Based on the improved polar representation, we propose a two-stream backbone and a novel neck, which is shown in Figure~\ref{fig3: structure_SiamPolar}. Furthermore, we introduce the details of three key components: asymmetric Siamese network, repeated cross correlation, and semi-FPN.

\begin{figure*}
	\centering
	\includegraphics[width=0.84\linewidth]{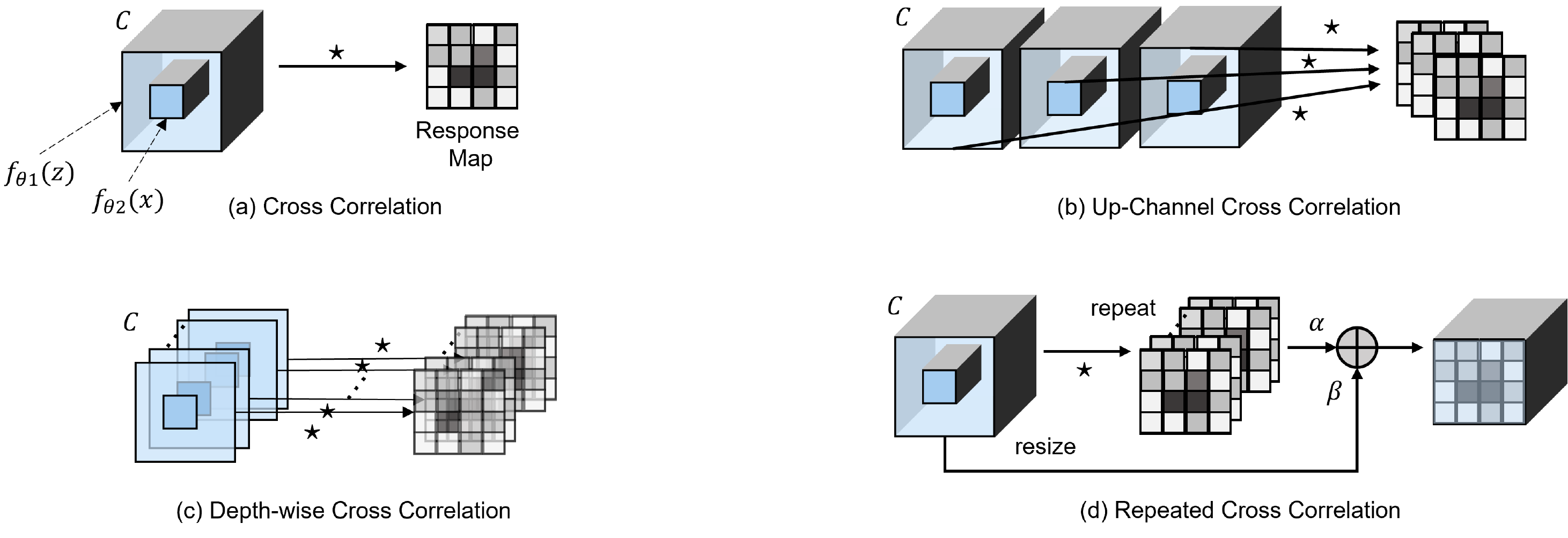}
	\caption{Comparison of the correlation method. (a) Cross correlation. (b) Up-channel cross correlation. (c) Depth-wise cross correlation. (d) Repeated cross correlation. $f_{\theta 1} (z)$ and $f_{\theta 2} (x)$ denote the search feature and template feature. $\star$ represents the convolution operation. $C$ is the number of channel. $\alpha$ and $\beta$ are the two trainable parameters.}
	\label{fig10: re_cross_correlation}
\end{figure*}

\begin{figure}
	\includegraphics[width=\linewidth]{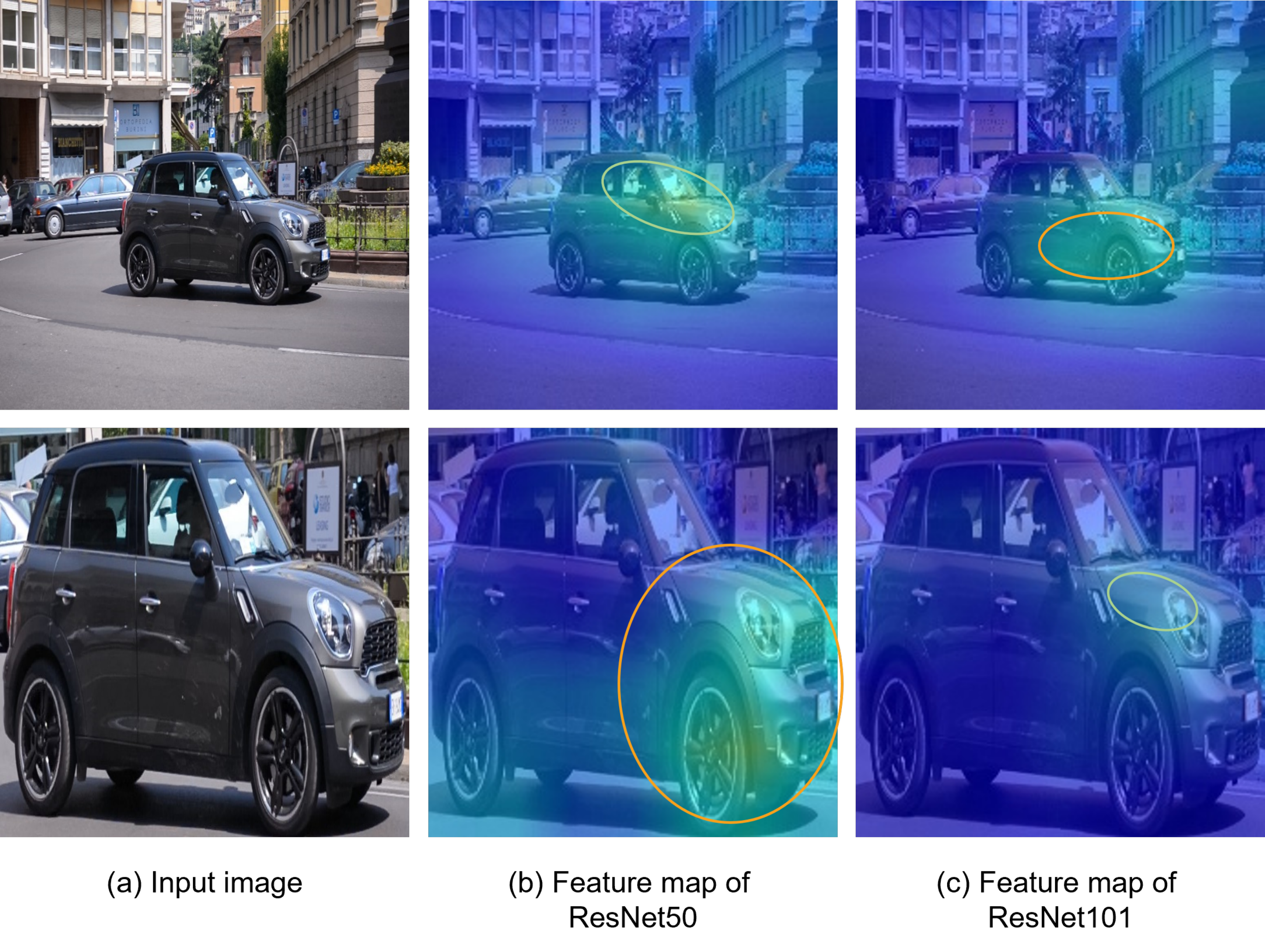}
	\caption{The visualization of feature maps extracted by ResNet50 and ResNet101. In the upper row of (a), the input is the search image, and the lower row of (a) is the template image. }
	\label{fig4: feature_vis}
\end{figure}

\subsection{Asymmetric Siamese Network}

Through experiments, it is found that deeper backbones lead to unaligned spatial problems between the search branch and the template branch. As shown in Figure~\ref{fig4: feature_vis}, gradCAM \cite{selvaraju2017grad} is applied to visualize the image features. These features are generated by feating the search image and the template image into a classification network (ResNet50 pretrained on the ImageNet dataset). The feature maps of ResNet101 in template image and search image don't focus on the same position. The former pay attention to the front of the car (including the wheels), while the latter only pay attention to the vicinity of the front lights. However, the feature map of ResNet50 highlights the front of the car. Hence, the features with orange circles are chosen in our network. 

In the neck of the network, the correlation similarity is computed by
\begin{equation}
	g(z, x) = f_{\theta 1}(z) \star f_{\theta 2}(x)
\end{equation}
where $z$ and $s$ are the search image and the template image.
 $f_{\theta 1}$ and $f_{\theta 2}$ are the two backbones. In the previous Siamese networks, $f_{\theta 1} \equiv f_{\theta 2}$. The template features are considered as a filter to locate the most similar region in the search features of the search image. If the two features focus on the different locations, the correlation similarity cannot be computed accurately. We define it as the spatial mis-alignment problem. 

The asymmetric Siamese network is proposed to improve the spatial misalignment problem. The network uses two different depth backbones (ResNet50 and ResNet101) rather than the two same depth backbones. ResNet50 is utilized in the template branch, while ResNet101 is applied to the search branch. These two backbones can extract the features from different size images at the same level. 

Through the asymmetric Siamese network, the deeper networks are introduced into the VOS tasks. The network can obtain higher-level semantic information, while the segmentation results are more accurate. The spatial misalignment problem is solved directly and efficiently. 

\subsection{Neck with Peeling Convolutions}\label{peeling_convolution}

As introduced in Section~\ref{intro}, the conflict between classification and regression tasks is a well-known problem \cite{peng2017large, song2020revisiting}. Thus the decoupled head for classification and localization is widely used in the most of detectors in both single image \cite{lin2017focal, tian2019fcos} and video \cite{li2018high, wang2019fast}. However, the features in the other part of the network has not been decoupled, e.g. feature pyramid network and cross-correlation. To further decoupled the features, we proposed the idea of peeling convolution. Based on that, a novel neck of the network is designed. 

The most used correlation computation methods include cross correlation \cite{bertinetto2016fully}, up-channel cross correlation \cite{li2018high} and depth-wise cross correlation \cite{girshick2014rich}, but these computations are designed for object tracking rather than object segmentation. The detailed structure is shown in Figure~\ref{fig10: re_cross_correlation}. Besides, the dense convolution increases the computation cost. These problems can be both reasoned to the similar regions lack detailed information for segmentation. Therefore, we repeat the single-channel similarity and fuse it with the original search features, which meets the design idea of peeling convolutions. Thus, the outputs of repeated cross correlation are more conducive to the VOS tasks. 

A light version is provided that the feature pyramid network (FPN) is not utilized in the network. It shows the advantage of polar representation in speed. However, it is found that semi-PFN can raise the accuracy without much speed loss. In FPN, the search features of different scales are integrated by interpolation, pixel-by-pixel addition and convolution (\texttt{FPN\_conv}). 
The integrated features are then forwarded to the branches of the polar head. In our method, the \texttt{FPN\_conv} is removed to get the semi-FPN without convolutions (Figure~\ref{fig7: semi_fpn}). In fact, the complete FPN is embedded in semi-FPN and polar head for there are stacked convolutions in the polar head replacing \texttt{FPN\_conv}. The experimental results demonstrate that semi-FPN shows a significant positive effect for the VOS tasks. 

\section{Experiments}

The proposed SiamPolar is established on MMdetection framework \cite{chen2019mmdetection} and we use NVIDIA RTX 2080 Ti GPU for all the experiments. When training, the number of rays is set to 36, which is an expirimental number. The discussion about the influence of rays' number is took in Section~\ref{ray_number_influence}. For backbones, we load the pre-trained ResNet, which is provided by MMdetection officially. Correlation similarity is calculated on the output of \texttt{conv5\_x}. In polar head, the scale ranges are set to \texttt{(-1, 256), (256, 512), (512, 1024), (1024, 2048), (2048, 1e8)}. 

\subsection{Improved Polar Representation}

\begin{figure*}
	\centering
	\includegraphics[width=\linewidth]{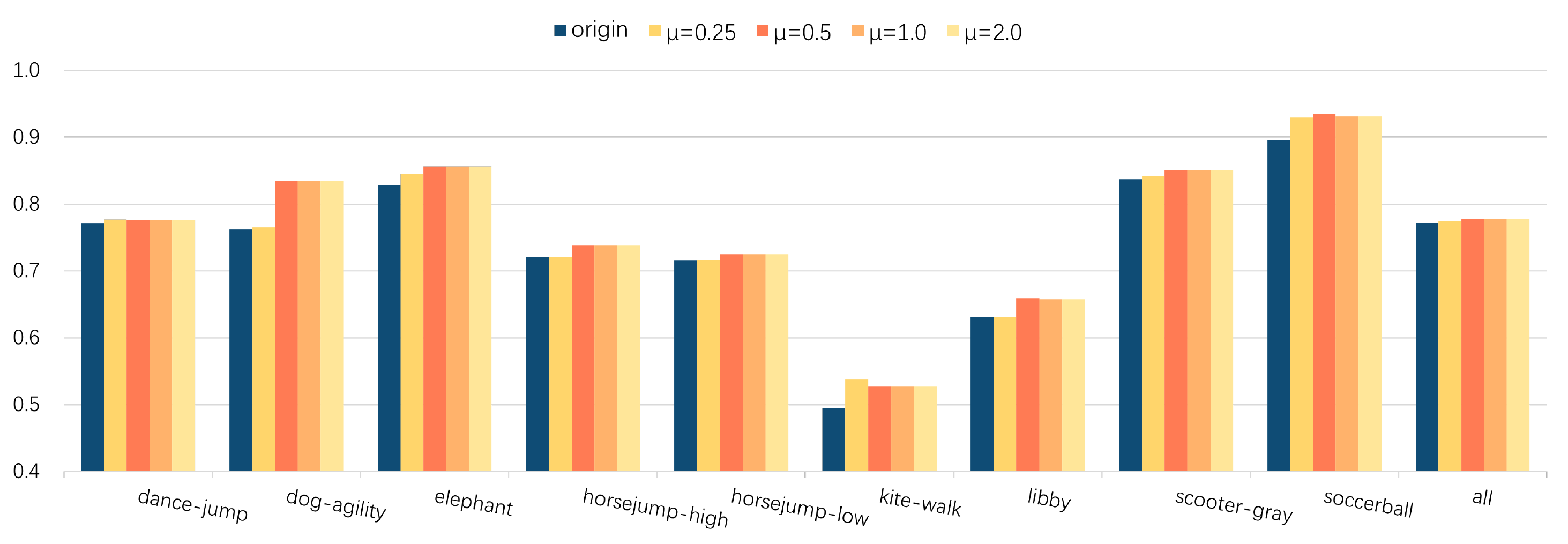}
	\caption{Intersection over union (IoU) scores between generated mask and ground truth. The abscissa is the name of each video clip. These are representative objects with complex contours. 'All' denotes the average IoU among videos. $\mu$ is a parameter in the algorithm of merging contours. }
	\label{fig11: generated_mask_all}
\end{figure*}

\begin{figure}
	\centering
	\includegraphics[width=\linewidth]{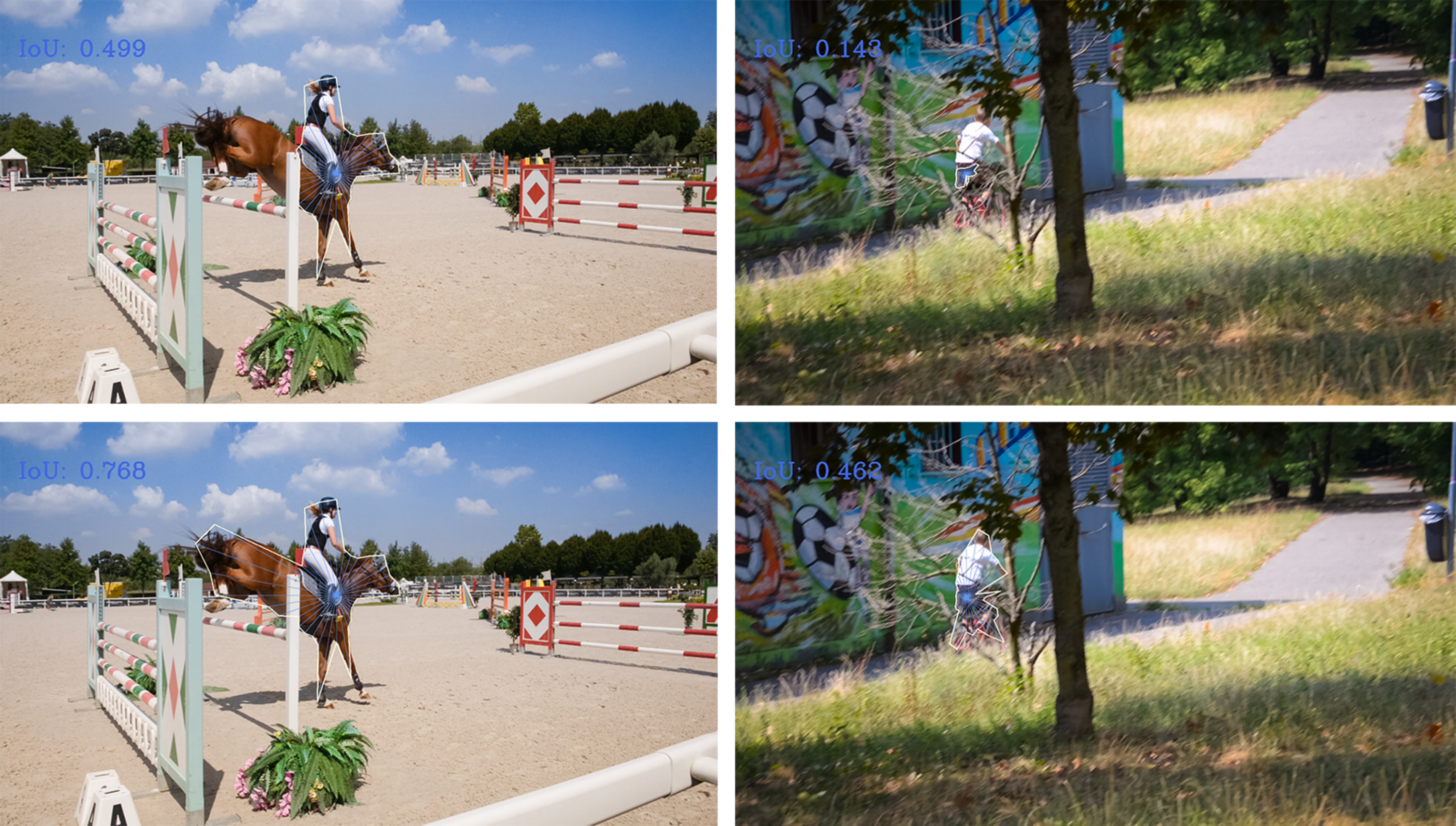}
	\caption{A visualization of generated labels (36 rays) on occluded objects. The first row is the masks generated by the original method in \cite{xie2020polarmask}, and the second row is the masks generated by the improved method. }
	\label{fig10: generated_mask}
\end{figure}

The masks assembled by generated labels are also the upper bound of the network. To verify that the improved polar representation performs better on the objects with complex contours, we compute the accuracy of each video (shown in Figure~\ref{fig11: generated_mask_all}). Since the position of the center candidate won't affect the accuracy, here the center is supposed to be the mass center to simplify the description. It shows that for different object the best $\mu$ is different. For all the videos $\mu = 0.5$ is the best choice, which gets the upper bound of $0.7784$. 

The visualization results are shown in Figure~\ref{fig10: generated_mask}. In addition to the improvement complex contoured objects, the representation accuracy of occluded objects has been significantly increased. 

\subsection{Performance on Benchmarks}

\begin{figure}[tpb]
	\centering
	\subfloat[drift \#21]{
		\begin{minipage}[t]{0.32\linewidth}
			\centering
			\includegraphics[width=1.06in]{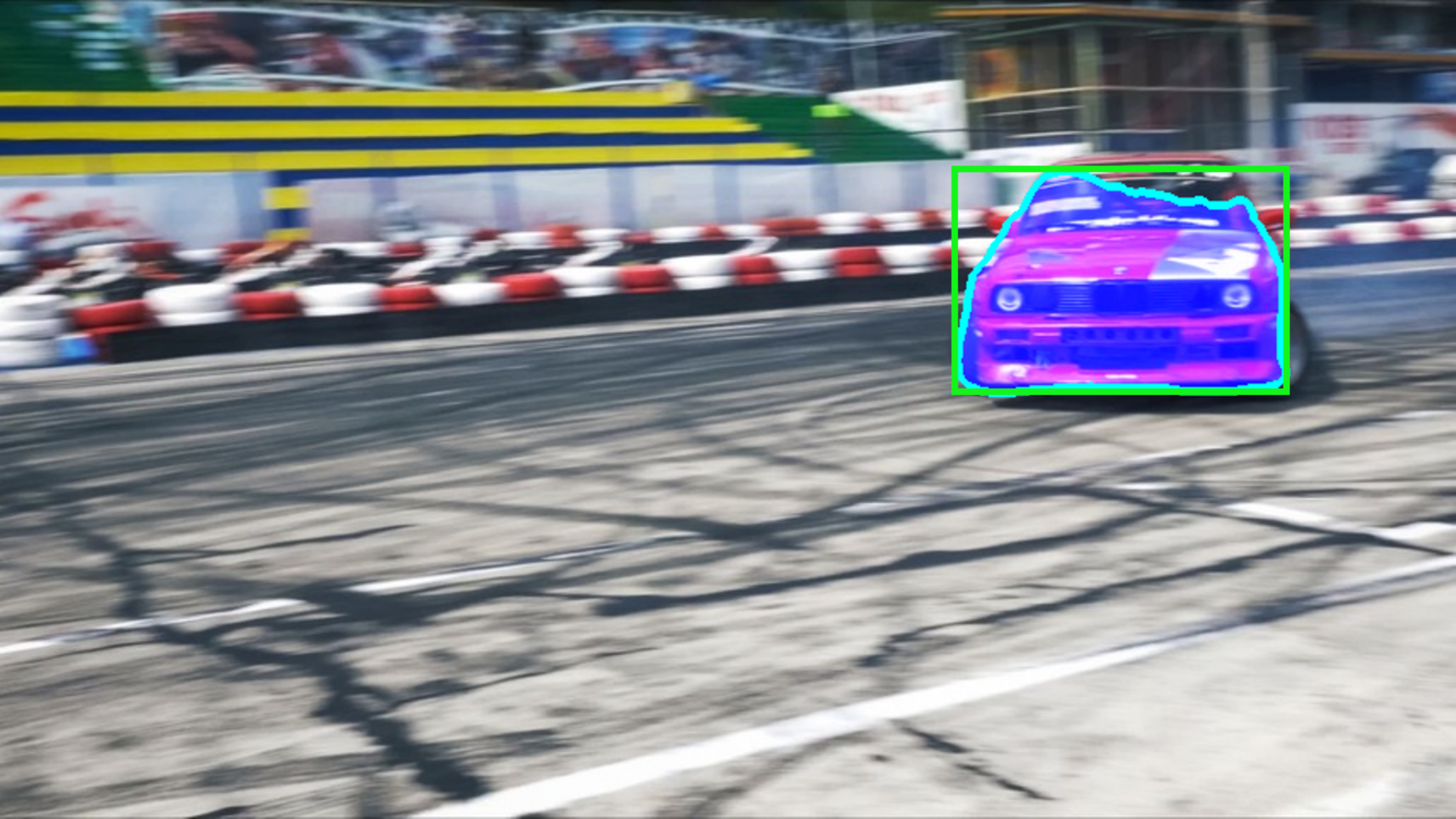}\\
			\vspace{0.05cm}
			\includegraphics[width=1.06in]{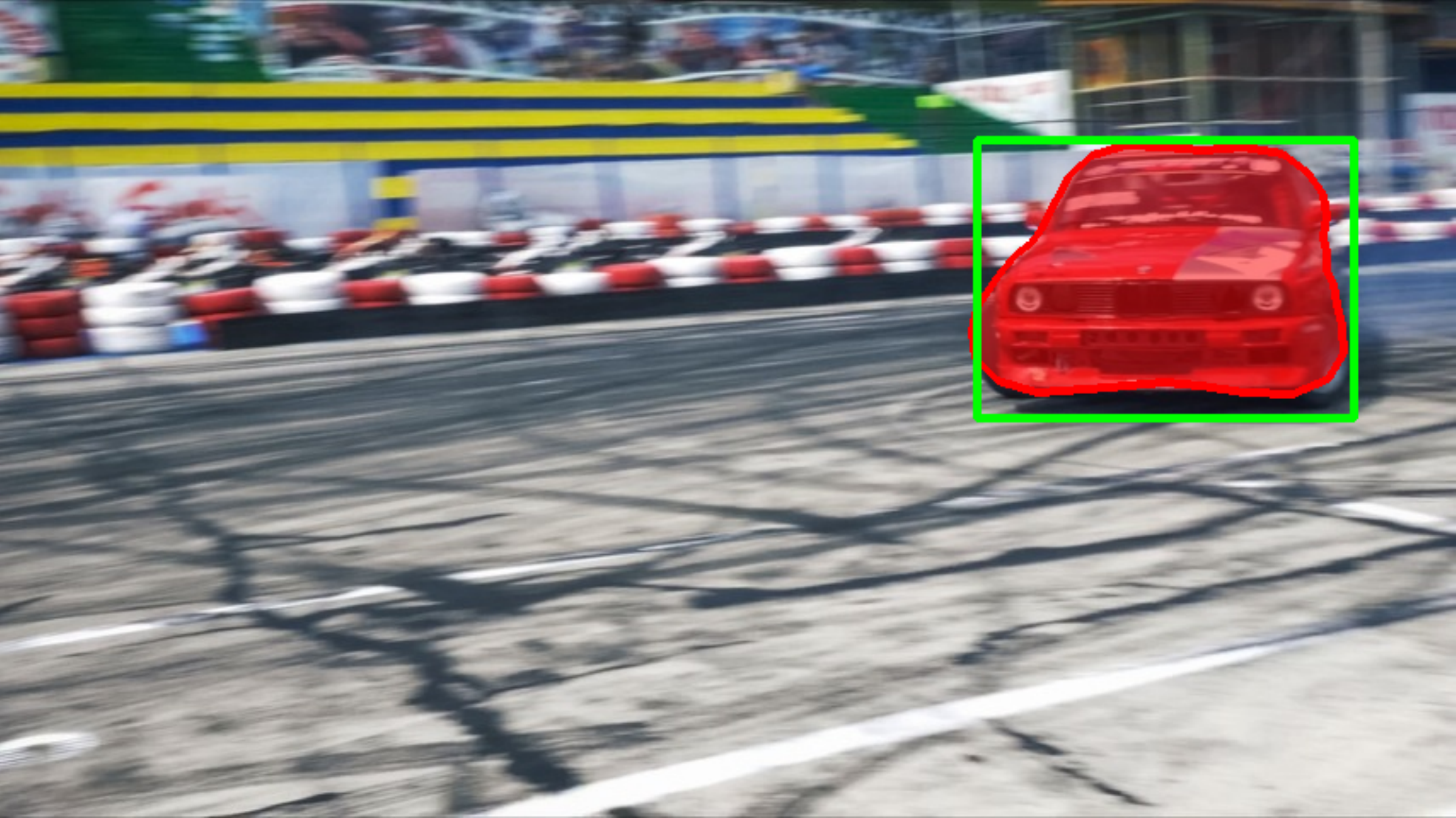}\\
		\end{minipage}%
	}
	\subfloat[drift \#31]{
		\begin{minipage}[t]{0.32\linewidth}
			\centering
			\includegraphics[width=1.06in]{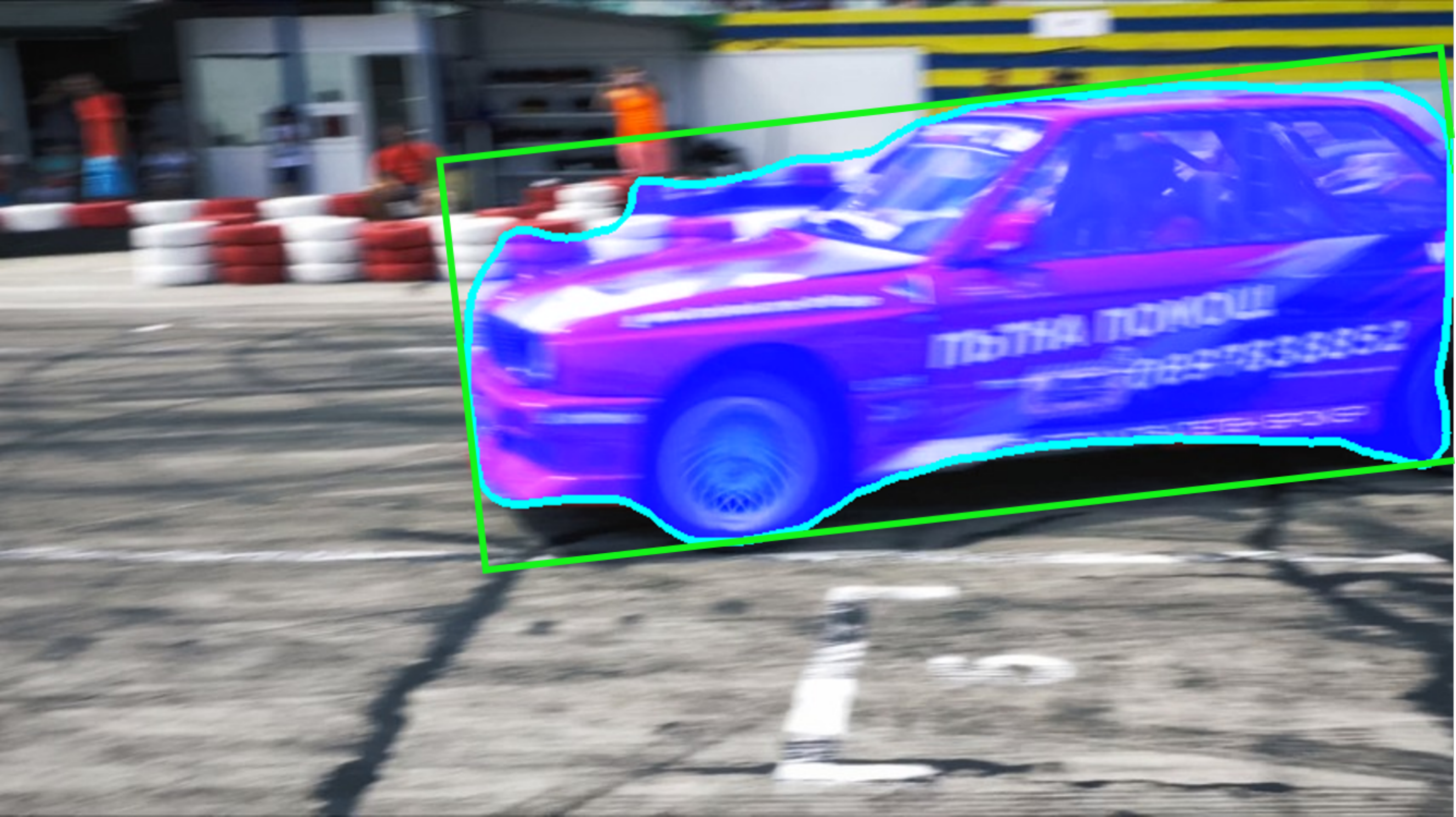}\\
			\vspace{0.05cm}
			\includegraphics[width=1.06in]{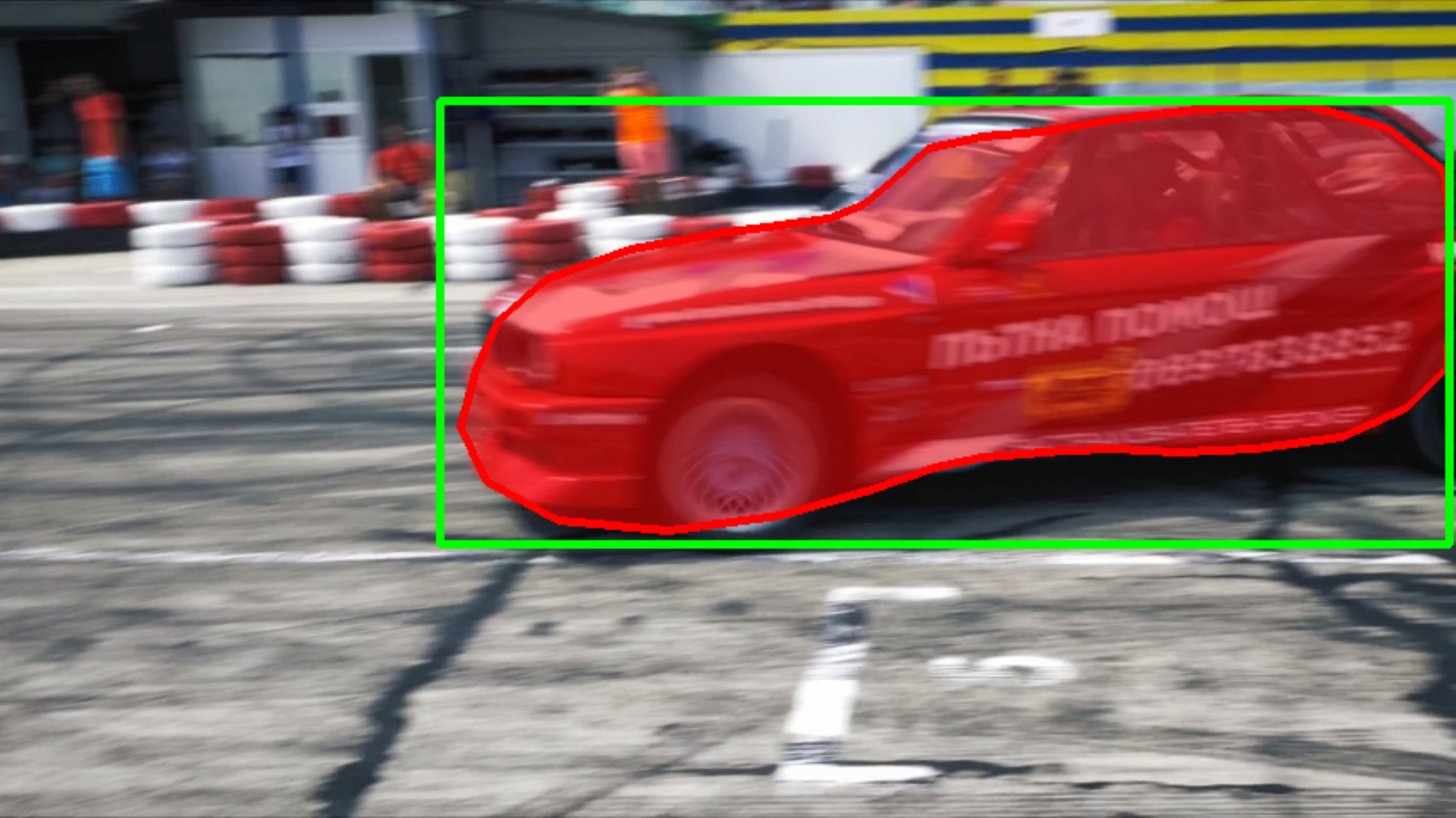}\\
		\end{minipage}%
	}
	\subfloat[paragliding \#62]{
		\begin{minipage}[t]{0.32\linewidth}
			\centering
			\includegraphics[width=1.06in]{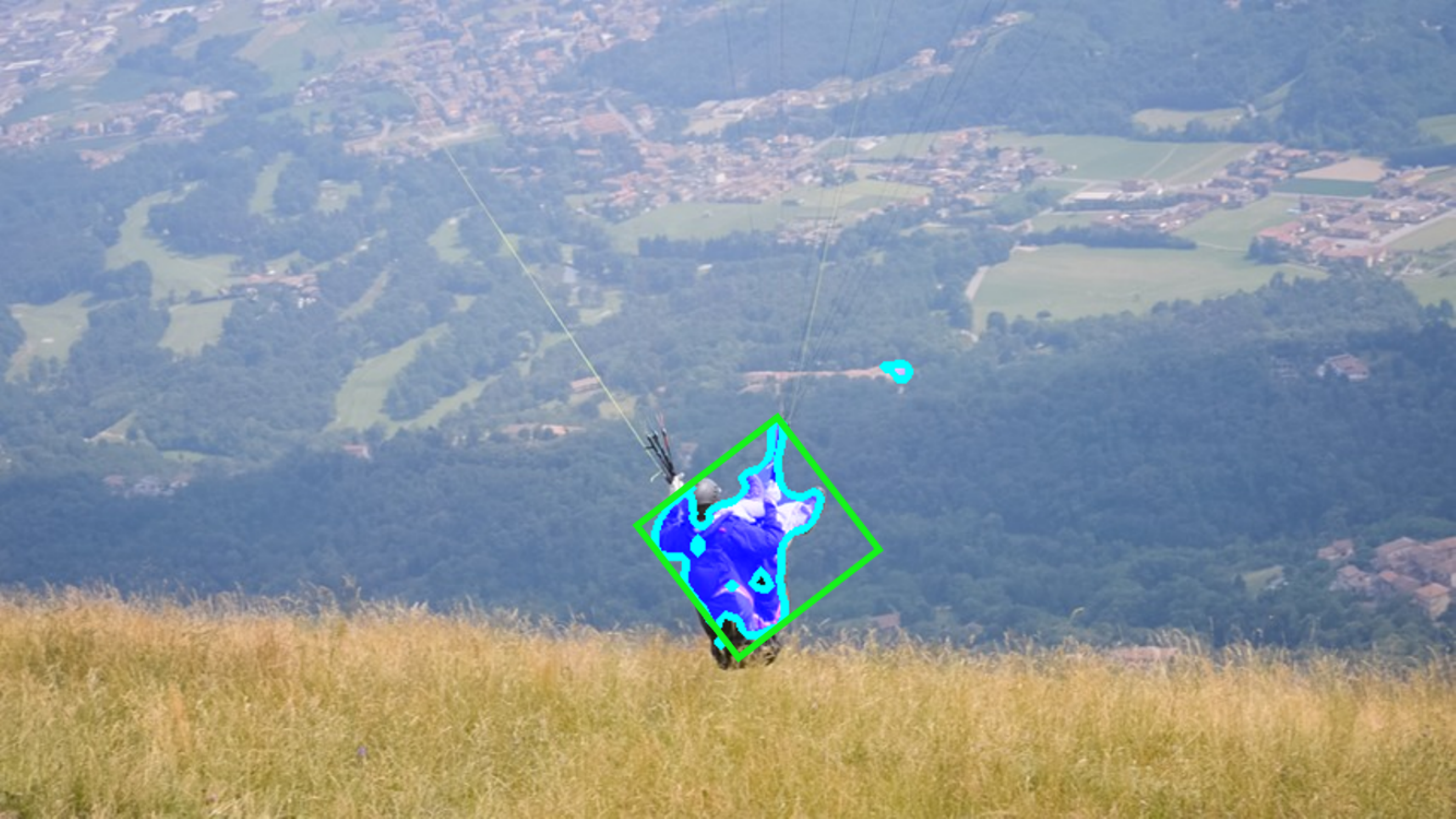}\\
			\vspace{0.05cm}
			\includegraphics[width=1.06in]{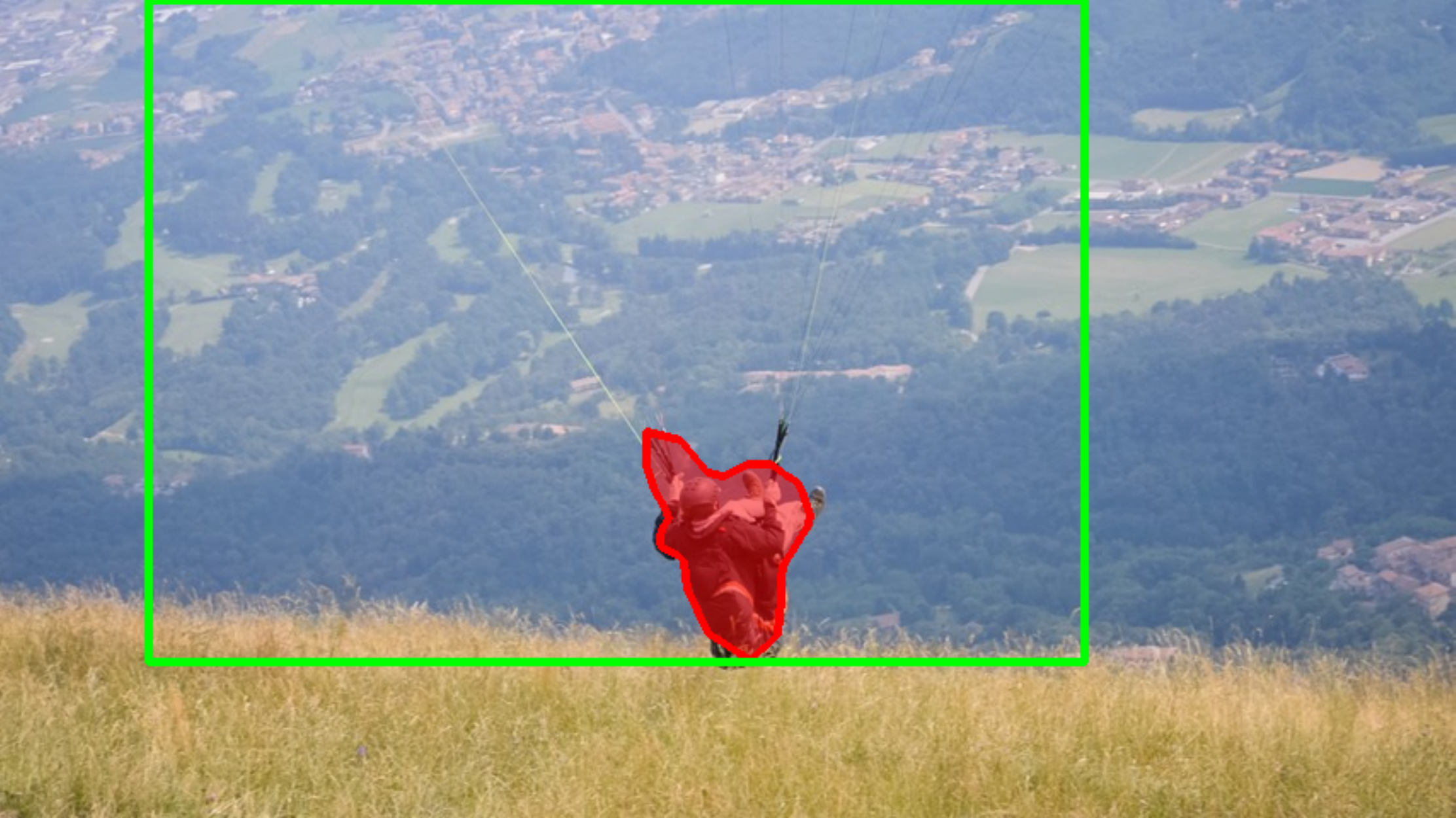}\\
		\end{minipage}%
	}
	\caption{The performance of real-time VOS methods, SiamMask \cite{wang2019fast} (blue) and SiamPolar (red) on DAVIS2016 dataset. SiamPolar effectively relieve the fragmentized edges of the objects, resulting in a smoother contour. The green boxes are the results of object tracking, and SiamMask uses rotated boxes.}
	\label{fig5: Compare SiamMask and SiamPolar}
\end{figure}

\begin{table}
    \footnotesize
    \centering
    \renewcommand
    \arraystretch{1.2}
    \caption{Quantitative comparison on DAVIS-2016 Val benchmark. F and M respectively denote if the method requires fine-tuning and whether it is initialised with a mask (\checkmark) or a bounding box (\xmark). Frame per second is used to measure the speed. }
    \label{table2}
    \begin{tabularx}{8.4cm}{p{1.3cm}|p{0.1cm}|p{0.1cm}|p{0.3cm}p{0.3cm}p{0.4cm}|p{0.3cm}p{0.3cm}p{0.4cm}|p{0.5cm}}
    \hline
    Method & F & M & $\mathcal J_M$ & $\mathcal J_R$ & $\mathcal J_D$ & $\mathcal F_M$ & $\mathcal F_R$ & $\mathcal F_D$ & Speed \\ \hline
    Lucid \cite{khoreva2019lucid} & \cmark & \cmark & {\bf 83.9} & 95.0 & 9.1 & 32.0 & 88.1 & 9.7 & 0.03 \\
    OSVOS \cite{caelles2017one} & \cmark & \cmark & 79.8 & 93.6 & 14.9 & {\bf 80.6} & 92.6 & 15.0 & 0.11 \\
    MSK \cite{perazzi2017learning} & \cmark & \cmark & 79.7 & 93.1 & 9.9 & 75.4 & 87.1 & 9.0 & 0.10 \\
    SFL \cite{cheng2017segflow} & \cmark & \cmark & 76.1 & 90.6 & 12.1 & 76.0 & 85.5 & 10.4 & 0.10 \\
    JMP \cite{fan2015jumpcut} & \cmark & \cmark & 57.0 & 62.6   & 39.4  & 53.1 & 54.2   & 38.4  & 0.08 \\ \hline
    FEELVOS \cite{voigtlaender2019feelvos} & \xmark & \cmark & 81.1 & 90.5   & 13.7  & 32.2 & 36.6   & 14.1  & 1.96  \\ 
    RGMP \cite{wug2018fast} & \xmark & \cmark & 81.5 & 91.7   & 10.9  & 32.0 & 90.8   & 10.1  & 8.00  \\ 
    PML \cite{chen2018blazingly} & \xmark & \cmark & 75.5 & 39.6   & 8.5   & 79.3 & {\bf 93.4} & 7.3   & 3.60  \\
    OSMN \cite{yang2018efficient} & \xmark & \cmark & 74.0 & 37.6   & 9.0   & 72.9 & 84.0   & 10.6  & 7.14  \\
    PLM \cite{shin2017pixel} & \xmark & \cmark & 70.2 & 36.3   & 11.2  & 62.5 & 73.2   & 14.7  & 6.70  \\
    VPN \cite{jampani2017video} & \xmark & \cmark & 70.2 & 32.3   & 12.4  & 65.5 & 69.0   & 14.4  & 1.59  \\
    OFL & \xmark & \cmark & 68.0 & 75.6   & 26.4  & 63.4 & 70.4   & 27.2  & 0.02  \\ \hline
    SiamMask \cite{wang2019fast} & \xmark & \xmark & 71.3 & 86.8   & 3.0   & 67.8 & 79.8   & {\bf 2.1} & 55.00 \\ 
    BVS \cite{marki2016bilateral} & \xmark & \xmark & 60.0 & 66.9   & 28.9  & 58.8 & 67.9   & 21.3  & 1.19  \\ 
    {\bf Ours-light} & \xmark & \xmark & 66.4 & 89.7 & {\bf 0.2} & 50.8 & 52.2 & 0.4  & {\bf 63.20} \\ 
    {\bf Ours} & \xmark & \xmark & 71.6 & {\bf 95.8} & 0.8 & 56.8 & 61.2   & 17.2  & 59.20 \\ \hline
    \end{tabularx}
\end{table}

\begin{table}
    \footnotesize
    \centering
    \renewcommand
    \arraystretch{1.2}
    \caption{Quantitative comparison on DAVIS-2017 Val benchmark. F and M respectively denote if the method requires fine-tuning and whether it is initialised with a mask (\checkmark) or a bounding box (\xmark). Frame per second is used to measure the speed.}
    \label{table7}
    \begin{tabularx}{8.4cm}{p{1.5cm}|p{0.2cm}|p{0.2cm}|p{0.6cm}p{0.6cm}|p{0.6cm}p{0.6cm}|p{0.6cm}}
    \hline
    Method & F & M & $\mathcal J_M$ & $\mathcal J_R$ & $\mathcal F_M$ & $\mathcal F_R$ & Speed \\ \hline
    OnAVOS & \cmark & \cmark & 61.6 & 67.4 & 69.1 & 75.4 & 0.10 \\
    OSVOS \cite{caelles2017one} & \cmark & \cmark & 56.6 & 63.8 & 63.9 & 73.8 & 0.10 \\ \hline
    FAVOS & \xmark & \cmark & 54.6 & 61.1 & 61.8 & 72.3 & 0.80 \\
    OSMN \cite{yang2018efficient} & \xmark & \cmark & 52.5 & 60.9 & 57.1 & 66.1 & 8.00 \\ \hline
    SiamMask & \xmark & \xmark & 54.3 & 62.8 & 58.5 & 67.5 & 55.00 \\
    Ours & \xmark & \xmark & 55.2 & 69.4 & 39.0 & 23.7 & 59.20 \\ \hline
    \end{tabularx}
\end{table}

\begin{table}
    \footnotesize
    \centering
    \renewcommand
    \arraystretch{1.2}
    \caption{Average pixel error (APE) scores on the SegTrack dataset (lower scores mean better). Note that it is unreasonable to simply average the APE across all sequences as the object size is a sensitive factor, thus “Average” is not computed.}
    \label{table5}
    \begin{tabularx}{8.4cm}{l|p{0.8cm}p{0.8cm}p{0.8cm}p{0.8cm}p{0.8cm}}
    \hline
         & MM \cite{chen2019multilevel} & HVS \cite{grundmann2010efficient} & TVS \cite{wang2016optimal} & HOP \cite{hao2020higher} & Ours \\ \hline
        birdfall & 201 & 305 & 163 & 232 & {\bf 102} \\ 
        cheetah & 856 & 1219 & {\bf 688} & 1032 & 818 \\ 
        girl & {\bf 987} & 5777 & 1186 & 1075 & 7231 \\ 
        monkeydog & 388 & 493 & 354 & 441 & {\bf 334} \\ 
        parachute & 223 & 1202 & {\bf 209} & 229 & 466 \\ 
        penguin & 661 & 2116 & 456 & 1254 & {\bf 387} \\ \hline
        Runtime (s/f) & 6.9s &  0.28s &  – & 0.53s & {\bf 0.19s} \\ \hline
    \end{tabularx}
\end{table}

\begin{table}
    \footnotesize
    \centering
    \renewcommand
    \arraystretch{1.2}
    \caption{Intersection over union (IoU) scores on the SegTrack v2 dataset. \#1 denotes the first object in that video.}
    \label{table4}
    \begin{tabularx}{8.4cm}{l|p{0.8cm}p{0.8cm}p{0.8cm}p{0.8cm}p{0.8cm}}
    \hline
         & FGS \cite{li2013video} & MM \cite{chen2019multilevel} & TS \cite{xiao2016track} & JOTS \cite{wen2015jots} & Ours \\ \hline
        bmx-person & 85.4  & 78.6  & 86.1  & {\bf 88.9}  &  61.6 \\ 
        dritf\#1 & 74.8  & 85.2  & 70.7  & 67.3 & {\bf 86.3} \\ 
        frog & 72.3  & 75.1  & {\bf 80.2}  & 56.3 & 56.7 \\ 
        hummingbird\#1 & 54.4  & 60.7  & 53.0 & 58.3 & {\bf 61.8} \\ 
        monkey & 84.8  & 80.4  & 83.1  & {\bf 86.0} & 77.4 \\ 
        soldier & {\bf 83.8}  & 77.3  & 76.3  & 81.1 & 70.4 \\ 
        worm & 82.8  & 81.3  & 82.4  & 79.3  & {\bf 82.5} \\ 
        Average & {\bf 76.9}  & {\bf 76.9}  & 75.9  & 73.8  & 72.8 \\ \hline
        Runtime (s/f) & 80s  & 14.9s  & 5.8s  & 20s &  {\bf 0.2s} \\ \hline
    \end{tabularx}
\end{table}

\begin{figure*}
	\centering
	\includegraphics[width=\linewidth]{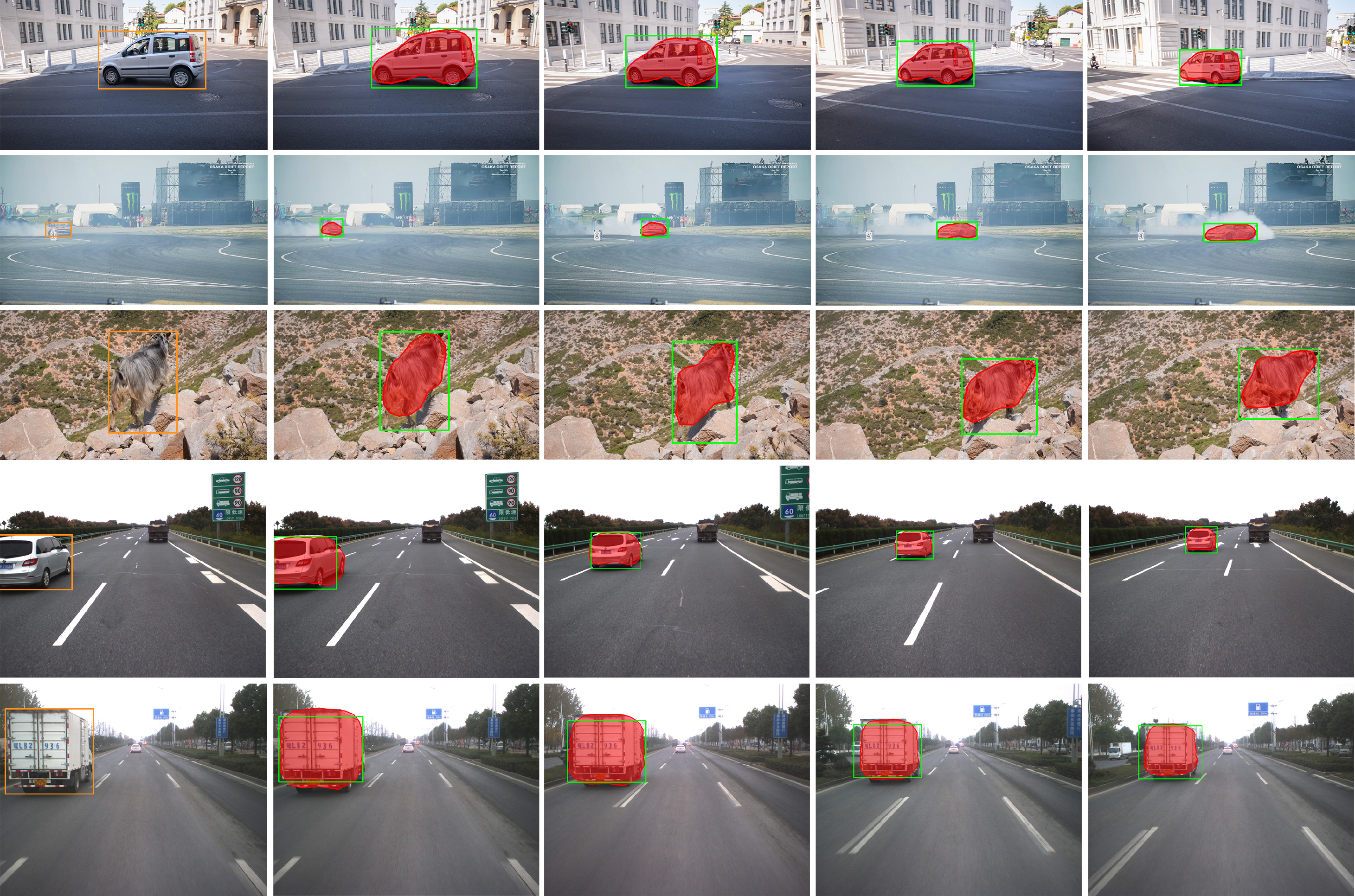}
	\caption{Qualitative results on the DAVIS2016 dataset and real traffic scenes TSD-max dataset. The first three rows are from the DAVIS2016 dataset and the last two rows are from the TSD-max dataset. The first column is the inputs, which are images and initial bounding boxes (shown in orange) rather than the initial masks.}
	\label{fig6: performance}
\end{figure*}

SegTrack \cite{TsaiBMVC10} and SegTrack v2 \cite{FliICCV2013} are commonly used to validate the video object segmentation algorithms. The datasets contain 6 and 14 dense annotated videos respectively. Besides, the most recently challenging benchmark is DAVIS2016 \cite{7780454} and DAVIS2017 \cite{Pont-Tuset_arXiv_2017}. They comprise a total of 50 sequences (30 videos for training and 20 for validation), all captured at 24fps and full HD 1080p spatial resolution. In each video sequence of DAVIS2016, a single instance is annotated. 
The DAVIS2017 dataset contains multiple instances in each video. 
The SiamPolar model is conducted multiple times in each video to segment the objects independently. 
All these datasets  cover several challenges in the VOS task, such as background clutter, fast motion, scale validation, camera shake, etc. 

The evaluation results compared with the benchmarks are shown in Table~\ref{table2}, Table~\ref{table7}, Table~\ref{table5} and Table~\ref{table4} respectively. For DAVIS2016 and DAVIS2017 dataset, we use the official performance measures \cite{khoreva2017lucid}: the Jaccard index ($\mathcal{J}$) to evince region similarity and the contour accuracy ($\mathcal{F}$) to evince contour accuracy. They could be computed by the following formulas: 
\begin{equation}
	\mathcal{J} = \frac{|M \cap G|}{|M \cup G|}
\end{equation}
where $M$ denotes the output mask and $G$ denotes the ground-truth mask. The output mask $M$ can also be interpreted as a set of closed contours $c(M)$ delimiting the spatial extent of the mask. Then we can compute the contour-based precision and recall $P_c$ and $R_c$ between the contour points of $c(M)$ and $c(G)$. 
\begin{equation}
	\mathcal{F} = \frac{2P_cR_c}{P_c+R_c}
\end{equation}
For each measure, mean $\mathcal{M}$, recall $\mathcal{R}$, and decay $\mathcal{D}$ are considered, which informs us about the gain or loss of performance over time. In order to be consistent with the benchmarks of the SegTrack datasets. We use intersection over union (IoU) and average pixel error (APE) scores to express the segmentation performance. IoU is the same as the Jaccard index and average pixel error is defined as: 
\begin{equation}
	APE = \frac{1}{N} \times \sum_{i=0}^{N} |M \backslash G|
\end{equation}
where $N$ is the number of input images. 

Most semi-supervised video object segmentation methods require initialized binary mask \cite{marki2016bilateral, fan2015jumpcut, jampani2017video, shin2017pixel}, fine-tuning \cite{bao2018cnn, maninis2018video, siam2019video}, data augmentation \cite{khoreva2017lucid, li2018video}, optical flow \cite{bao2018cnn, cheng2018fast, li2018video} and other computationally intensive technologies, so it is difficult to achieve real-time speed. 
The experimental results in Table~\ref{table2} demonstrate that our method is effective for real-time video object segmentation. Both of the light versions of SiamPolar and SiamMask never use FPN. Moreover, the light version of SiamPolar is 8.2 fps faster than the SiamMask method. 

In addition to the speed metric, Table~\ref{table2} and Table~\ref{table7} also show that SiamPolar reaches higher accuracy than real-time VOS methods. It is equitable to compare our method with SiamMask, even though SiamMask doesn't equip asymmetric backbones. Because SiamMask solves the spatial mis-alignment problem by padding the inputs and keeping the objects the same size among inputs, so that the objects are in the same vision level. 
Compare to SiamMask, our method is more stable in edges. As a result, our method is more suitable for a real scene, as shown in Figure~\ref{fig5: Compare SiamMask and SiamPolar}. In a real traffic scene, objects blurry because of the high speed of the vehicles. For computer vision tasks, inaccurate contours of obstacles are fatal in the real scene, which may lead to serious accidents. 
Besides, the bounding box branch and the mask branch are separated in both SiamMask and SiamPolar.
As a result, the segmentation result may not be consistent with the edge of the tracking result. Figure~\ref{fig5: Compare SiamMask and SiamPolar} (c) shows that SiamPolar tracks more details than SiamMask, such as the ropes of the paragliding. 
It is also demonstrated that SiamPolar achieves low decay and high recall \cite{7780454} for region similarity. 
This indicates that the proposed method performs well in different categories and  is robust over time. 
Moreover, it is a very effective video segmentation method that does not require fine-tuning. 
The visualization results are shown in Figure~\ref{fig6: performance}. 
The proposed SiamPolar method performs stable on the VOS task as only the bounding boxes are needed.

On SegTrack and SegTrack v2 dataset SiamPolar performs a excellente speed, and it make great accuracy in most categories.  However, for the objects which have complex contours, it is difficult to generate accurate polar represented mask. In that situation, the methods based on pixels may get a higher scores than the contours based method. For instance, multi-modal mean (MM) \cite{chen2019multilevel} performances better on the "girl" sequence in SegTrack dataset than ours. In SegTrack v2, Joint online tracking and segmentation (JOTS) \cite{wen2015jots} is better on the "bmx-person", "soldier" sequence. 

\subsection{Ablation Studies}

\begin{table*}[width=\textwidth]
    \renewcommand{\arraystretch}{1.2}
    \caption{Results of Ablation Studies on DAVIS-2016. PC: polar centerness, AB: asymmetric backbone. \cmark means the improvement is added to the model. '-' means the improvement is not added to the model. }
    \label{table1}
    \begin{tabular*}{\tblwidth}{cc|ccc|ccc|ccc|c}
    \hline
    Backbone & Neck & Contours Merger & improved PC & AB & 
    {$\mathcal J_M$} & {$\mathcal J_R$} & {$\mathcal J_D$} & {$\mathcal F_M$} & {$\mathcal F_R$} & {$\mathcal F_D$} & Speed (fps) \\ \hline
    ResNet101 & Connection & \cmark & \cmark & \cmark & 66.4 & 89.7 & 0.2 & 50.8 & 52.2 & 0.4 & 63.20 \\
    ResNet101 & semi-FPN & \cmark & \cmark & \cmark & 71.6 & 95.8 & 0.8 & 56.8 & 61.2   & 17.2  & 59.20 \\ 
    ResNet101 & semi-FPN & - & \cmark & \cmark & 70.2 & 93.9 & 0.4 & 55.6 & 56.7 & 19.6 & 59.20 \\ 
    ResNet101 & semi-FPN & - & - & \cmark & 69.1 & 91.4 & -0.5 & 53.2 & 2.0 & 11.8 & 59.20 \\
    ResNet101 & FPN & - & - & \cmark & 68.9 & 91.8 & -1.5 & 52.6 & 54.1 & 15.3 & 37.80 \\
    ResNet50 & FPN & - & - & - & 60.5 & 77.3 & 1.6 & 44.0 & 35.3 & 13.9 & 33.20 \\
    ResNet101 & FPN & - & - & - & 53.3 & 65.8 & -2.5 & 36.2 & 23.8 & 15.5 & 26.40 \\ \hline
    \end{tabular*}
\end{table*}

\begin{table}
    \footnotesize
    \centering
    \renewcommand
    \arraystretch{1.2}
    \caption{Performances with different stride region size of candidate center on DAVIS2016.}
    \label{table8}
    \begin{tabularx}{8.4cm}{l|p{0.7cm}p{0.7cm}p{0.7cm}|p{0.7cm}p{0.7cm}p{0.7cm}}
    \hline
    Stride size & $\mathcal J_M$ & $\mathcal J_R$ & $\mathcal J_D$ & $\mathcal F_M$ & $\mathcal F_R$ & $\mathcal F_D$ \\ \hline
    1.0 & 69.4 & 95.4 & 1.2 & 51.9 & 46.6 & 13.5 \\
    1.5 & 71.6 & 95.8 & 0.8 & 56.8 & 61.2 & 17.2 \\
    2.0 & 71.0 & 96.0 & 1.5 & 55.6 & 54.4 & 10.2 \\ \hline
    \end{tabularx}
\end{table}

\textbf{Stride Size: }The stride size of center candidate is examined, as shown in Table~\ref{table8}. 
The performance gets improved as the stride size increases from $1.0$ to $1.5$. 
However, the performance gets worse when the stride size keeps increasing from $1.5$ to $2.0$. 
When the stride size equal to or larger than $2.5$, the network does not converge for the center candidates fall outside the masks. 
Hence, we set the stride size to  $1.5$ in the experiments.

\textbf{Improved Polar Representation: }With the $1.5$ stride size, an ablation study is conducted to evaluate the improved polar coordinate representation and the effectiveness of each module, as shown in Table~\ref{table1}. The contours merger and the new Polar Centerness improve the detection accuracy as introduced in Section~\ref{improve_polar_representation}.

\textbf{Asymmetric Backbones: }When the depth of the backbone network increases from ResNet50 to ResNet101, the regional similarity drops from 60.5\% to 53.3\% because of the unaligned spatial problem. The asymmetric Siamese network helps to improve the accuracy to 68.9\%. Moreover, due to the limited depth types of ResNet, we only enumerate one kind of asymmetric Siamese network. The depth combination of ResNet50 and ResNet101 may not be the optimal deep combination scheme. In our work, we aim to provide new ideas for the design of Siamese Networks. More efficient two stream backbone can be designed in the future. 

\textbf{Neck: }The model with a directly connected neck is also the light version of SiamPolar. Compare to the model with semi-FPN and FPN neck, it gets the highest speed. The semi-FPN method reduces common convolutions in different branches of polar head, reducing the antagonism of their features. More importantly, removing convolutions help to reduce the network parameters, so the speed of our model has been greatly improved from 37.8fps to 59.2fps. 

\begin{figure}[tb]
	\centering
	\includegraphics[width=\linewidth]{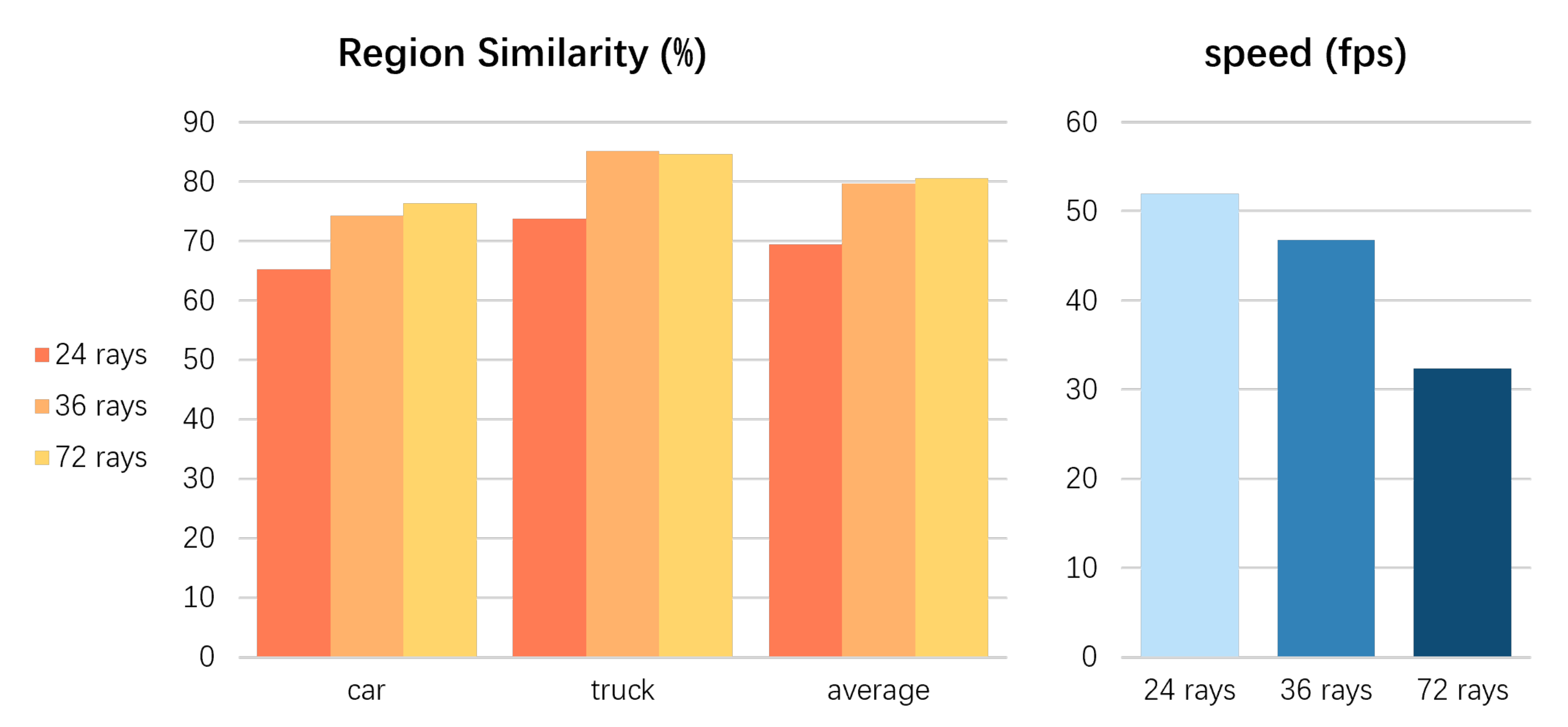}
	\caption{Accuracy and speed performance on real traffic scene TSD-max dataset. The influences of the number of rays are shown. }
	\label{fig9: tsd_max}
\end{figure}

\subsection{Application in Traffic Scene}\label{ray_number_influence}

\begin{figure}
	\centering
	\includegraphics[width=\linewidth]{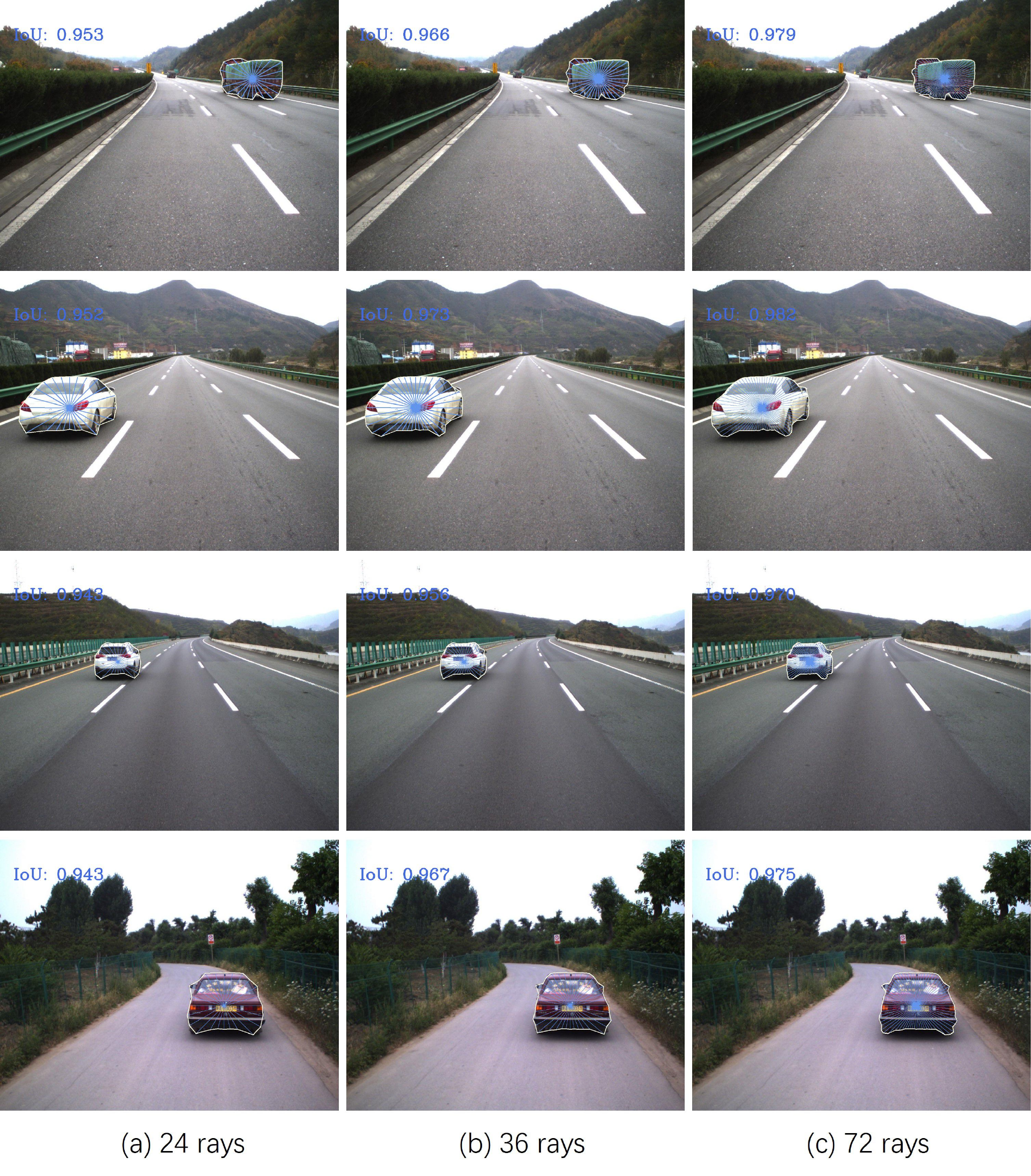}
	\caption{A visulization of generated masks on TSD-max dataset with different number of rays. IoU between the generated mask and the ground-truth mask is labeled on the left top corner of each figure, which is also the upper bound of SiamPolar.}
	\label{fig8: polar_number}
\end{figure}

In the real traffic scenes, there are environmental interference like  backlight, the objects are blocked and the objects are deformed, etc. In order to evaluate the robustness, we test SiamPolar on a traffic scenes dataset, namely TSD-max. It is a real traffic scenes dataset, which is constructed by the Institution of Artificial Intelligence and Robostics, Xi'an Jiaotong University. The dataset includes 11 manually labeled videos (6 car videos, 3 truck videos and 2 bus videos), where one car sequence and one truck sequence are kept as validation dataset and the others are used to train. 

Besides, we further discuss the effect of the polar numbers on SiamPolar (Figure~\ref{fig9: tsd_max}). In the edge point generation algorithm, we apply fixed angle points to simulate the mask, which sacrifices the representation accuracy. Visualization results of polar representation are shown in Figure~\ref{fig8: polar_number}. The gap between ground-truth and polar represented masks comes from two reasons. One reason for the greater impact is, if the number of rays is too few, the protruding part of objects will be cut. Besides, if there are two contour points on one ray, we will choose the remoter one as polar representation. So if the number of rays is too many, the recessed part may be incorrectly filled. 

The results indicate that the best polar number for VOS task in the traffic scene is 72. Because 72 rays' model reaches 80.5\% region similarity on average. However, as the number of rays increases, the network parameters will increase and the inference speed will decrease. We make SiamPolar flexible to automatically choose polar numbers for different applications. Therefore, SiamPolar can be used in different scenes. 

\section{Conclusions}

The SiamPolar method proposed in this paper is a real-time video object segmentation method using Siamese network. To the best of our knowledge, this is the first work to introduce polar coordinate mask into VOS tasks, which greatly improved the speed of VOS. In the SiamPolar framework, we propose three contributions to solve the problem of spatial alignment and improve the efficiency of search features: the asymmetric Siamese network, semi-FPN, and repeated cross correlation. Without fine-tuning and initializing the mask, it is demonstrated that SiamPolar is fast and accurate compared with other methods. Moreover, the idea of peeling convolutions to reduce antagonism can be utilized for future Siamese Networks design.

\bibliographystyle{cas-model2-names}

\bibliography{cas-dc-template}

\end{document}